\documentclass[conference]{IEEEtran}
\IEEEoverridecommandlockouts
\usepackage{cite}
\usepackage{amsmath,amssymb,amsfonts}
\usepackage{algorithmic}
\usepackage{graphicx}
\usepackage{textcomp}
\usepackage{xcolor}
\usepackage{multirow}
\usepackage{hyperref}
\def\BibTeX{{\rm B\kern-.05em{\sc i\kern-.025em b}\kern-.08em
    T\kern-.1667em\lower.7ex\hbox{E}\kern-.125emX}}
\begin{document}

\title{
    AlignKT: Explicitly Modeling Knowledge State for Knowledge Tracing with Ideal State Alignment
    \thanks{\IEEEauthorrefmark{2}~Corresponding author.}
    \thanks{\IEEEauthorrefmark{1}~Jing Xiao and Chang You contributed equally. This work is supported by the National Nature Science Foundation of China (Project No. 62177015).}
}

\author{
  \IEEEauthorblockN{Jing Xiao\IEEEauthorrefmark{1}\IEEEauthorrefmark{2}}
  \IEEEauthorblockA{
    \textit{School of Computer Science} \\
    \textit{South China Normal University} \\
    Guangzhou, China \\
    xiaojing@scnu.edu.cn
  }
  \and
  \IEEEauthorblockN{Chang You\IEEEauthorrefmark{1}}
  \IEEEauthorblockA{
    \textit{School of Artificial Intelligence} \\
    \textit{South China Normal University} \\
    Foshan, China \\
    2023025161@m.scnu.edu.cn
  }
  \and
  \IEEEauthorblockN{Zhiyu Chen}
  \IEEEauthorblockA{
    \textit{School of Computer Science} \\
    \textit{South China Normal University} \\
    Guangzhou, China \\
    2023010230@m.scnu.edu.cn
  }
}

\maketitle

\begin{abstract}
Knowledge Tracing (KT) serves as a fundamental component of Intelligent Tutoring Systems (ITS), enabling these systems to monitor and understand learners' progress by modeling their knowledge state. However, many existing KT models primarily focus on fitting the sequences of learners' interactions, and often overlook the knowledge state itself. This limitation leads to reduced interpretability and insufficient instructional support from the ITS. To address this challenge, we propose AlignKT, which employs a frontend-to-backend architecture to explicitly model a stable knowledge state. In this approach, the preliminary knowledge state is aligned with an additional criterion. Specifically, we define an ideal knowledge state based on pedagogical theories as the alignment criterion, providing a foundation for interpretability. We utilize five encoders to implement this set-up, and incorporate a contrastive learning module to enhance the robustness of the alignment process. Through extensive experiments, AlignKT demonstrates superior performance, outperforming seven KT baselines on three real-world datasets. It achieves state-of-the-art results on two of these datasets and exhibits competitive performance on the third. The code of this work is available at https://github.com/SCNU203/AlignKT.
\end{abstract}


\section{Introduction}
\label{sec:introction}

Today, online education generates a large amount of educational data on intelligent tutoring systems, knowledge tracing is precisely the important tool for ITS to perceive learners’ learning progress, and its function is to assess learners’ future performance with these educational data. We first give some definitions of the concepts related to the KT task.
\emph{Knowledge Concepts} are a set of knowledge, skills, and other elements designated by experts in relevant disciplines that learners need to master. \emph{Interaction Sequence} represents the learners' historical exercise records on ITS. \emph{Knowledge State} represents the abstract summary of learners' mastery level of each Knowledge Concept.
\begin{figure}[htbp]
\centerline{\includegraphics{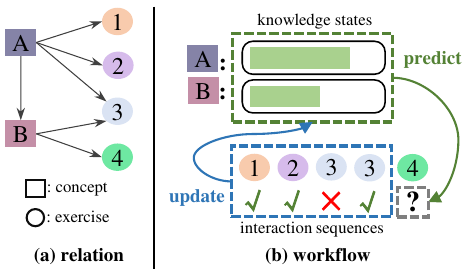}}
\caption{A simplified workflow diagram of DLKT. (a) depicts a small set of knowledge concepts alongside their corresponding exercises, with directed arrows indicating the relation among them. (b) illustrates the process by which DLKT model updates knowledge state and predicts next response.}
\label{fig-intro}
\end{figure}
Fig.\ref{fig-intro} shows the general process of deep learning based knowledge tracing (DLKT) for assessing learners' future performance, where (a) depicts the dependency structure in an example: Exercise 1 covers Concept A, and Concept B is also dependent on Concept A. In Fig.\ref{fig-intro}(b), DLKT model updates learners' knowledge state from their responses and predicts subsequent exercise performance.

However, updating a learner's knowledge state in real-world educational scenarios is not a straightforward task. Knowledge state, as an internally latent and not directly measurable state of learners, lacks a theoretically grounded and intuitive representation. Previous work solely utilized DLKT as a prediction tool, and designed the model training paradigm with a focus on fitting learners' interaction sequences. This approach inherently restricts the model to learning only the observable patterns of data variation within the sequences, while neglecting the modeling of the underlying knowledge state itself.
Even worse, due to the influence of some instability factors, such as the sparsity of learners' interaction data (learners do not fully engage with all concepts and complete all exercises, with the number of exercises far exceeding the number of concepts) and the fluctuation of knowledge state (this is more common in the case of short sequences, especially when students are disturbed by random events such as carelessness), this further complicates the modeling process that develops a long-term stable and interpretable representation of the knowledge state.

On one hand, the coexistence of the aforementioned two issues in practical application jointly constrains the ability of KT models to measure knowledge state; on the other hand, as an abstraction of learners' mastery level of knowledge concepts, knowledge state is crucial for the interpretability of KT models: ITS necessitates high interpretable feedback from it to ensure the effectiveness of its guidance.
This poses a dilemma. To facilitate subsequent research, we propose the following Research Objectives (RO) based on the two issues:

\begin{itemize}
\item RO1: Explicitly modeling knowledge state with a high interpretable paradigm.
\item RO2: Improve the prediction performance, based on RO1.
\end{itemize}

Centering around the research objectives, we propose an interpretable modular model AlignKT. AlignKT comprises three modules: a frontend for modeling preliminary knowledge state, a backend for aligning the state with an additional criterion, and a contrastive learning module for enhancing the robustness of this process. The first two modules are integrated to form the frontend-to-backend architecture, which is designed to accomplish the RO1. We use the \emph{ideal state} as the alignment criterion. The ideal state is that learners possess a thorough mastery of each knowledge concept, which is regarded as hand-crafted feature defined by the discipline experts. 
AlignKT constructs an interpretable representation of the \emph{ideal knowledge state} based on this premise. In the alignment, AlignKT compares the knowledge state with the ideal knowledge state, and incorporates the interpretable representation from latter into former. This is a source of the AlignKT's interpretability.

From the perspective of model-design to accomplish the ROs, we also implement the following key innovations: 
\begin{itemize}
\item To alleviate data sparsity, we modify the \emph{Rasch Model-based Embedding} (denoted as \emph{M-RME}), which accounts for individual variations across exercises to enhance the information contained within the representations.
\item To improve the interpretabiliy of the model, we design a \emph{Time-and-Content Balanced Attention} (denoted as \emph{TCBA}) method, which integrates insights from both pedagogical theory and cognitive psychology to model learners' forgetting behavior effectively.
\item To mitigate the fluctuation of knowledge state, we employ \emph{contrastive learning module} (denoted as \emph{CL}) that improves model's robustness by enhancing its ability to discriminate minor different semantics.
\end{itemize}
These advancements have enabled AlignKT to achieve state-of-the-art performance on several real-world datasets, and all specific methodological details are elaborated in the subsequent sections.

\section{Related Work}
\label{sec:relatedwork}
\subsection{Knowledge Tracing}
The seminal work that introduced deep learning to knowledge tracing is DKT\cite{DKT_2015}. DKT treated the hidden state as the knowledge state, and utilized it to predict learners' responses. DKT did not explicitly model the knowledge state.

SAKT\cite{SAKT_2019} was the first to apply attention mechanism\cite{attention_2023} to knowledge tracing. The attention-based DLKT model captures inter-concept relationships via attention scores and models learners’ knowledge state through these dependencies. AKT\cite{AKT_2020} leveraged the Rasch Model-based Embedding method to alleviate the issue of data sparsity, and achieved excellent prediction performance. FolibiKT \cite{folibiKT_2023} extended the AKT model by introducing a learnable linear bias term to simulate the forgetting behavior of learners. 

DKVMN\cite{DKVMN_2017} and DTransformer\cite{dTransformer_2023} used additional memory to explicitly model the knowledge state. 
However, both methods initialized their memory randomly and obtained the knowledge state during the training process. This approach failed to circumvent the black-box nature of deep learning, resulting in limited improvement of interpretability.

\subsection{Contrastive Learning}
Contrastive Learning is a self-supervised approach that fosters the model to learn more intrinsic and semantic representations by constructing pairs of positive and negative samples and discriminating them. Positive pairs denote semantically similar instances, while negative pairs represent dissimilar or unrelated counterparts. Contrastive learning optimizes representation similarity by attracting positive pairs and repelling negative pairs. In this way, the model can discriminate the semantics between representations more clearly, thus mitigating the perturbations caused by instability factors.

Inspired by the SimCSE\cite{SimCSE_2021}, CL4KT\cite{CL4KT_2022} designed a method for generating positive and negative samples through data augmentation, and applied this method to AKT, significantly improving its prediction performance.

\section{The AlignKT Method}
\label{sec:method}
\subsection{Problem Statement}\label{3.1}
Knowledge tracing can be regarded as a supervised temporal sequence problem, and the deep learning knowledge tracing model considers the interaction sequence as a learning process. The interaction sequence up to the current time step $t$ is denoted as follows:
\begin{equation}
\label{eq-1}
    I_t=\{(e_1, c_1, r_1), ..., (e_t, c_t, r_t)\}, \quad e_t\in \mathbb{Q}, c_t\in \mathbb{C}, 
\end{equation}
where $\mathbb{Q}$ denotes the set of exercises, $\mathbb{C}$ denotes the set of knowledge concepts; $e_t$, $c_t$ respectively denote the integer index of exercise and knowledge concept; $r_t=0$ indicates an incorrect response, while $r_t=1$ indicates a correct one. Given the $I_{t-1}$ and $(e_t, c_t)$, DLKT model needs to predict the learner's next response $\hat{r}_t$.

\subsection{Modified Rasch Model-based Embedding}\label{3.2}
Given that the number of the knowledge concept $N_c$ in real-world scenarios is significantly smaller than the number of the exercise $N_e$, in order to avoid overfitting, previous KT models primarily used $c_i$ rather than $(e_i, c_i)$ to model the knowledge state. However, distinct exercises also contain significant information. Rasch model-based embedding\cite{AKT_2020} is utilized to capture individual differences among exercises covering the same concept, and the differences are treated as supplementary information and added to the embedding representation. Inspired by this, we modified the method for a further exploitation.

Firstly, we define the \emph{state} indicating whether a learner has mastered the concept $i$ as $s_i = c_i+N_c \cdot r_i$. When $r_i=0$, it indicates that the learner has not mastered the knowledge concept; when $r_i=1$, it indicates that the learner has mastered the concept. Secondly, we convert the $c_i$ and $s_i$ into one-hot vectors, subsequently map them into a $d$-dimensional representation space and get the embedding $\mathbf{c} \in \mathbf{R}^d$ and $\mathbf{s} \in \mathbf{R}^d$. The modified Rasch model-based embedding is as follows:
\begin{equation}
\label{eq-2}
    \mathbf{c_t}=(1-a1) \cdot \mathbf{c_t} + a1 \cdot [\mu_{e_t}(\mathbf{d_{c_t}}+\mathbf{f_{diff}}(\mu_{e_t}))], 
\end{equation}
\begin{equation}
\label{eq-3}
    \mathbf{s_t}=(1-a2) \cdot \mathbf{s_t} + a2 \cdot [\mu_{e_t}(\mathbf{d_{s_t}}+\mathbf{f_{diff}}(\mu_{e_t}))], 
\end{equation}
where $t$ denotes the current time step; $\mathbf{c_t}$ and $\mathbf{s_t}$ denote the embedding representation; $\mathbf{d_{c_t}} \in \mathbf{R}^d$ and $\mathbf{d_{s_t}} \in \mathbf{R}^d$ denote the summary of variations $e_t$ covering $c_t$ and $s_t$ respectively; $\mu_{e_t}$ denotes a scalar that reflect the difficulty of the $e_t$; the $\mathbf{f_{diff}}$ denotes a single linear layer that maps the $\mu_{e_t}$ to a $d$-dimensional embedding. 
Taking into account the variations in the ratio between $N_c$ and $N_e$ across different datasets, we introduce $a1$ and $a2$ to control the proportion of supplementary information in the $\mathbf{c_t}$ and $\mathbf{s_t}$, see \ref{subsec: imHyper} for details.

\subsection{Architecture}\label{3.3}
\subsubsection{The Frontend}
\begin{figure}[htbp]
\centerline{\includegraphics[width=\linewidth]{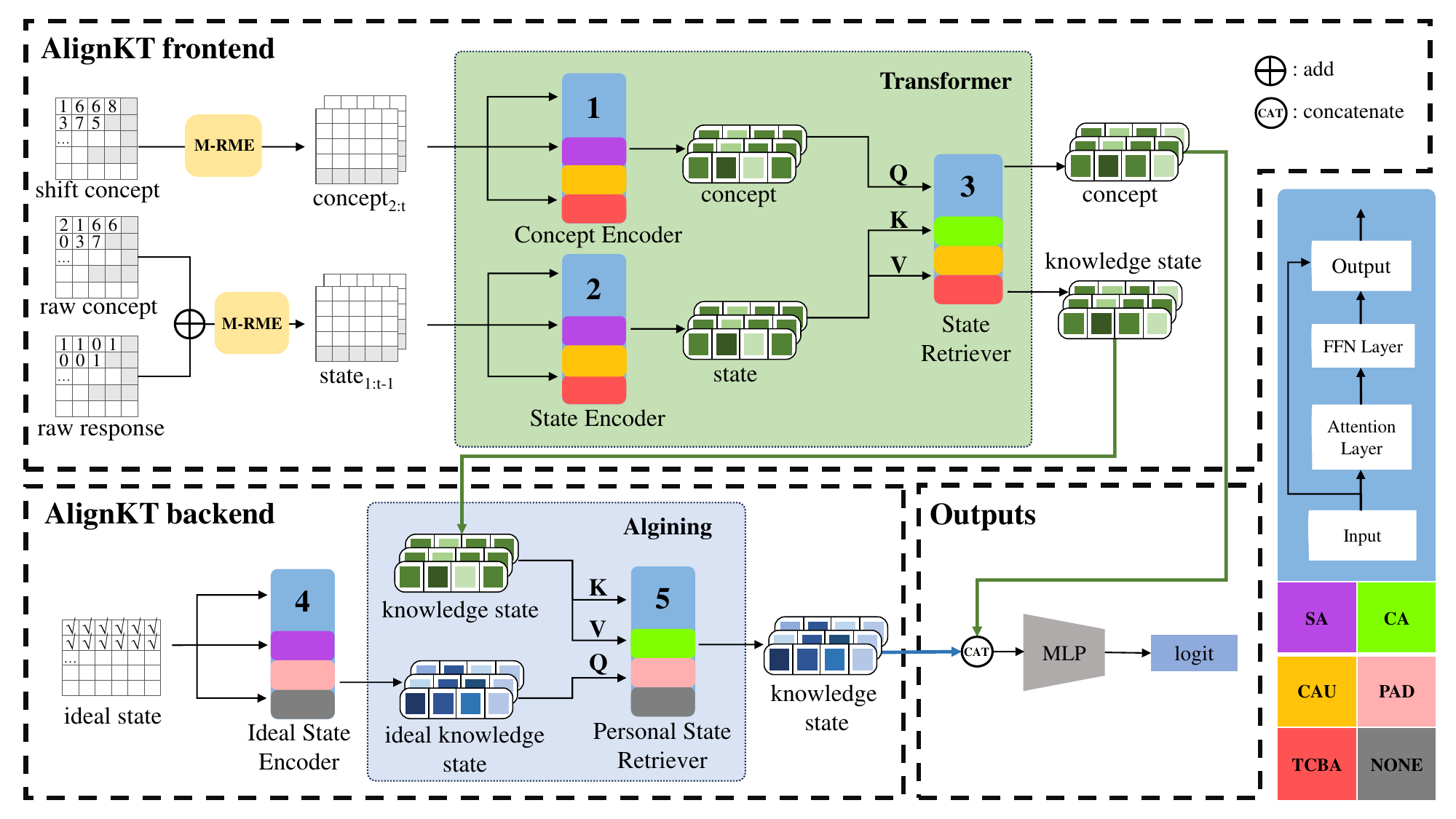}}
\caption{Overview of the AlignKT. The right sub-diagram shows the generic structure of the encoder, with the color blocks below representing different configurations. SA denotes Self-Attention; CA denotes Cross-Attention; CAU denotes a Causal Mask, and PAD denotes a Padding Mask. TCBA denotes Time-and-Content Balanced Attention, while NONE denotes the scaled dot-product attention. For the sake of overall brevity, the contrastive learning module is not depicted in the diagram.}
\label{fig-AlignKT}
\end{figure}

Fig.\ref{fig-AlignKT} shows the the overall workflow of AlignKT. 
AlignKT features five encoders configured diversely, three frontend encoders form a Transformer\cite{attention_2023} for temporal sequences, while the two backend encoders construct the ideal knowledge state, and then align the preliminary knowledge state from the frontend with it.

Frontend encoders use the self-attention mechanism to capture the relation within the input sequence, such as the dependent relation among knowledge concepts, or the prerequisite relation in mastery level.
We denote these encoders as \emph{Concept Encoder}, \emph{State Encoder} and \emph{State Retriever} respectively. 
Concept Encoder tackles the $\mathbf{c_{2:t}}$ at time steps ranging from $2$ to $t$ (start at $1$). At each time step, the current representation of knowledge concept aggregates the information from preceding knowledge concepts:
\begin{equation}
\label{eq-4}
    \mathbf{c_{2:t}} = \mathbf{F_{CE}}(\mathbf{Q:c_{2:t}},\mathbf{K:c_{2:t}},\mathbf{V:c_{2:t}}; CAU),
\end{equation}
where $\mathbf{F_{CE}}$ denotes the Concept Encoder, the CAU denotes a causal mask. When the attention mechanism is utilized for processing temporal sequences, a causal mask is necessary to ensure that future time steps are invisible to the current one, thus preventing information leakage.

The State Encoder models the prerequisite relation in $\mathbf{s}$: 
\begin{equation}
\label{eq-5}
    \mathbf{s_{1:t-1}} = \mathbf{F_{SE}}(\mathbf{s_{1:t-1}},\mathbf{s_{1:t-1}},\mathbf{s_{1:t-1}}; CAU),
\end{equation}
where $\mathbf{F_{SE}}$ denotes the State Encoder, and the $\mathbf{s_{1:t-1}}$ denotes the \emph{preliminary knowledge state}.

The State Retriever uses a cross-attention mechanism to further tackle the $\mathbf{c_{2:t}}$ and $\mathbf{s_{1:t-1}}$, this mismatch between $\mathbf{c}$ and $\mathbf{s}$ follows the process that learners try to solve the current exercise based on their past experiences.
Distinct from previous work\cite{AKT_2020} , we use the $\mathbf{c}$ as the \emph{Query}, use the $\mathbf{s}$ as the \emph{Key} and \emph{Value}:
\begin{equation}
\label{eq-6}
    \mathbf{c_{2:t}} = \mathbf{F_{SR}}(\mathbf{Q:c_{2:t}},\mathbf{K:s_{1:t-1}},\mathbf{V:s_{1:t-1}}; CAU),
\end{equation}
where $\mathbf{F_{SR}}$ denotes the State Retriever. We finally obtain the processed $\mathbf{c_{2:t}}$ as the query results across time steps from $2$ to $t$.

Notably, we implement TCBA as the method for computing attention score in the frontend encoders, which is designed to model learners' forgetting behavior over time. The specific details are provided in the \ref{3.4}.

\subsubsection{The Backend}
In AlignKT, the primary function of the backend is alignment, and is implemented using two encoders. 
We define the \emph{ideal state} as $s^*_i = c_i+N_c \cdot 1$, the subscript here denotes the index of knowledge concepts. $s^*$ denotes the ideal state that all knowledge concepts are fully mastered, and its embedding $\mathbf{s^*}$ share the representation space with $\mathbf{s}$. To get the \emph{ideal knowledge state}, we have:
\begin{equation}
\label{eq-7}
    \mathbf{s^*_{1:N_c}} = \mathbf{F_{ISE}}(\mathbf{s^*_{1:N_c}},\mathbf{s^*_{1:N_c}},\mathbf{s^*_{1:N_c}}; PAD),
\end{equation}
where $\mathbf{F_{ISE}}$ denotes the fourth encoder as \emph{Ideal State Encoder}. Due to irrelevance to temporal information, $\mathbf{F_{ISE}}$ employs a standard scaled dot-product attention mechanism and padding mask. In this configuration, $\mathbf{F_{ISE}}$ can bidirectionally model the correlation in $\mathbf{s^*_{1:N_c}}$ by considering both preceding and subsequent contextual state, and obtain a representation that takes into account more global information.

The fifth encoder named \emph{Personal State Retriever} uses a cross-attention mechanism to align the preliminary knowledge state obtained from frontend with the ideal knowledge state.
Similar to the $\mathbf{F_{SR}}$, we finally obtain the $\mathbf{s}$ as the interpretable and intuitive representation of the knowledge state:
\begin{equation}
\label{eq-8}
    \mathbf{s_{1:t-1}} = \mathbf{F_{PSR}}(\mathbf{s_{1:t-1}},\mathbf{s^*_{1:N_c}},\mathbf{s^*_{1:N_c}}; PAD), 
\end{equation}
where $\mathbf{F_{PSR}}$ denotes \emph{Personal State Retriever}. The retriever calculates the attention score between the learner's knowledge state $\mathbf{s_{i}}$ at each time step, and each mastered knowledge concept $\mathbf{s^*_{j}}$ in the ideal knowledge state that incorporates contextual information. Subsequently, the retriever combines them using the score as weights, thereby completing the alignment of the knowledge state.

\subsection{Time-and-Content Balanced Attention}\label{3.4}
In real-world scenarios, learning is always influenced by memory: unfamiliar knowledge/skills are easily forgotten, whereas familiar ones are not. \emph{Time-and-Content Balanced Attention} method used to model forgetting behavior is inspired by the \emph{Ebbinghaus Forgetting Curve} \cite{Psychology_2020} in cognitive psychology. We utilize a fitting formula denoted as $R_t=\exp{(-\frac{t}{d})}$ aiming to trade-off the impact between temporal distance and mastery level on memory, where $t$ denotes the duration of time, $d$ denotes the retention coefficient and $R_t$ denotes the degree of memory retention.
We denote the TCBA score based on the scaled dot-product attention score $\alpha_{t, i}$\cite{attention_2023} as:
\begin{equation}
\label{eq-9}
    \alpha_{t, i}'=\alpha_{t, i} \cdot \exp (-\frac{sin(\frac{\left| t-i \right|}{L})}{\gamma \cdot \frac{\mathbf{q}_t^T\mathbf{k}_i}{\sqrt{d}}}), 
\end{equation}
where $\mathbf{q}_t^T\mathbf{k}_i$ denotes the inner product that reflects mastery level in TCBA method; $\left| t-i \right|$ denotes the temporal distance; $L$ denotes the manually set maximum memory capacity, beyond which the temporal distances are truncated to ensure that the value of the exponential term in \eqref{eq-9} is always within the $(0,1]$; $\gamma$ denotes a learnable parameter that regulate the significance of the overall weights. In a word, a higher mastery level corresponds to a higher weighted score, and vice versa; score gradually decline over time.

Besides, the TCBA method leverages temporal distance to represent relative positional information, which is essential for the attention mechanism that employs parallel computation.

\subsection{Contrastive Learning Module}\label{3.5}
We adopt contrastive learning to enhance the model's capability of discriminating representations of different semantics. Given the requirement that a model with strong generalization should be resistant to minor perturbations, we artificially introduce noise into the $c$ and $s$ by randomly swapping their order or masking a certain position with [MASK], as the method proposed in CL4KT\cite{CL4KT_2022}. Subsequently, we obtain the positive samples $\mathbf{c^{pos}}$ and $\mathbf{s^{pos}}$ in \eqref{eq-4} and \eqref{eq-5} respectively.
For negative samples, we reverse the response in the original sequence and obtain $\mathbf{c^{neg}}$ and $\mathbf{s^{neg}}$ with the same procedure.
In AlignKT, we employ Information Noise-Contrastive Estimation (InfoNCE)\cite{CPC_2019} as a simpler and more generalized approach for computing contrastive loss:
\begin{equation}
\begin{aligned}
\label{eq-10}
    l^{CL}_c=-\log \frac{\exp{(\frac{\mathbf{sim}(\mathbf{c}, \mathbf{c}^{pos})}{\tau}) }}{\exp{(\frac{\mathbf{sim}(\mathbf{c}, \mathbf{c}^{neg})}{\tau})}+\exp{(\frac{\mathbf{sim}(\mathbf{c}, \mathbf{c}^{pos})}{\tau}) }} \\
    l^{CL}_s=-\log \frac{\exp{(\frac{\mathbf{sim}(\mathbf{s}, \mathbf{s}^{pos})}{\tau}) }}{\exp{(\frac{\mathbf{sim}(\mathbf{s}, \mathbf{s}^{neg})}{\tau})}+\exp{(\frac{\mathbf{sim}(\mathbf{s}, \mathbf{s}^{pos})}{\tau}) }}, 
\end{aligned}
\end{equation}
\begin{table}[htbp]
\scriptsize
\caption{AUC and ACC results on AS09, AL05 and NIPS34. Best models are \textbf{bold}, second best models are \underline{underline}.}
\label{table-1}
\begin{center}
\begin{tabular}{ccccccc}
\hline
\multirow{2}{*}{Model} & \multicolumn{3}{c}{AUC} & \multicolumn{3}{c}{ACC} \\
\cline{2-7}
&AS09&AL05&NIPS34&AS09&AL05&NIPS34  \\
\hline
DKT\cite{DKT_2015}&0.7552&0.8149&0.7689&0.7258&0.8097&0.7036 \\
DKVMN\cite{DKVMN_2017}&0.7479&0.8065&0.7673&0.7199&0.8027&0.7016 \\
AKT\cite{AKT_2020}&\underline{0.7805}&0.8139&\underline{0.7999}&\underline{0.7380}&0.8050&\textbf{0.7296} \\
DTransformer\cite{dTransformer_2023}&0.7763&0.8188&0.7991&0.7315&0.8043&0.7292 \\
FolibiKT\cite{folibiKT_2023}&0.7770&0.8159&\textbf{0.8004}&0.7325&0.8035&\underline{0.7293} \\
simpleKT\cite{simpleKT_2023}&0.7724&0.8212&0.7994&0.7293&0.8069&0.7292 \\
stableKT\cite{stableKT_2024}&0.7766&\underline{0.8275}&0.7990&0.7318&\underline{0.8102}&0.7285 \\
\hline
AlignKT&\textbf{0.7857}&\textbf{0.8323}&0.7987&\textbf{0.7382}&\textbf{0.8125}&0.7283 \\
\hline
\end{tabular}
\end{center}
\end{table}
where $\tau$ is a temperature hyperparameter that adjusts the sensitivity to the distance between positive and negative samples, and $\mathbf{sim}$ denotes the cosine similarity.

\subsection{Model Training}\label{3.6}
We concatenate the $\mathbf{s_{1:t-1}}$ obtained in \eqref{eq-8} with the $\mathbf{c_{2:t}}$ obtained in \eqref{eq-6}, and use a two-layer fully-connected network (denoted as $\mathbf{MLP}$) and a sigmoid function (denoted as $\mathbf{\sigma}$) for prediction:
\begin{equation}
\label{eq-11}
     \mathbf{\hat{r}_{2:t}}=\mathbf{\sigma}(\mathbf{MLP}([\mathbf{s_{1:t-1}} \oplus \mathbf{c_{2:t}}])).
\end{equation}

For each response $\hat{r}_i$ in the prediction result $\mathbf{\hat{r}_{2:t}}$, we use a binary cross-entropy to compute the loss:
\begin{equation}
\label{eq-12}
     l^{BCE}=\sum_i(r_i log(\hat{r}_i) + (1-r_i) log(1-\hat{r}_i)).
\end{equation}

Finally, AlignKT optimizes the linear combination of these two loss functions: 
\begin{equation}
\label{eq-13}
     loss=l^{BCE}+\lambda \cdot (l^{CL}_c+l^{CL}_s),
\end{equation}
where $\lambda$ denotes a hyperparameter that regulates the significance of contrastive learning module.

\section{Experiments}
We conduct all experiments on the open-source KT platform pyKT\cite{pyKT_2023}, including the standardized data preprocessing method and the evaluation process that fitting closely with real-world educational scenarios. 
We evaluate our model on three real-world educational datasets described as follows: 

\begin{itemize}
\item ASSISTments2009\footnote{https://sites.google.com/site/assistmentsdata/home/2009-2010-assistment-data/skill-builder-data-2009-2010} (AS09): collected from the \emph{ASSISTments online education platform}, there are $123$ concepts and $17,737$ exercises in the dataset, and the concept-to-exercise ratio is $0.0069$, average learner interaction length of $72.29$.
\item Algebra2005\footnote{https://pslcdatashop.web.cmu.edu/KDDCup/} (AL05): collected from the \emph{KDD Cup EDM Challenge}, there are $112$ concepts and $173,113$ exercises in the dataset, and the concept-to-exercise ratio is $0.0006$, average learner interaction length of $186.55$.
\item NeurIPS2020 Education Challenge (NIPS34)\footnote{https://eedi.com/projects/neurips-education-challenge}: collected from the \emph{NeurIPS 2020 Education Challenge}, there are $57$ concepts and $948$ exercises in the dataset, and the concept-to-exercise ratio is $0.0601$, average learner interaction length of $147.68$.
\end{itemize}

\subsection{Overall Performance}
\label{ovalPerf}
We compare our AlignKT with the seven baseline DLKT models, all of them are implemented using pyKT. The results are presented in Table \ref{table-1}, and we have the following observations: (1) our model demonstrates superior performance on the AS09 and AL05 compared with the baselines, and outperforms the second best models by $0.52\%$ (compared with AKT on the AS09) and $0.48\%$ (compared with stableKT on the AL05) in AUC respectively. (2) Compared with FolibiKT, which employs forgetting-aware linear bias to simulate learners' forgetting behavior, our model equipped with TCBA method outperforms in terms of AUC and ACC on both AS09 and AL05. Compared with DKVMN and DTransformer, which employ additional memory to explicitly model the knowledge state, our model not only explicitly models the knowledge state, but also achieves superior prediction performance in the aforementioned aspects. This indicates that the innovations proposed in accordance with the research objectives are operating effectively. (3) On the NIPS34, although our model does not perform optimal performance, the gap between our model and the optimal AUC is merely $0.17\%$ (compared with FolibiKT), and the performance differences between our model and the latest models such as simpleKT and stableKT fluctuate within a range of $\pm0.0003$.

\subsection{Impact of hyperparameters}
\label{subsec: imHyper}
\begin{figure}[htbp]
\centerline{\includegraphics[width=\linewidth]{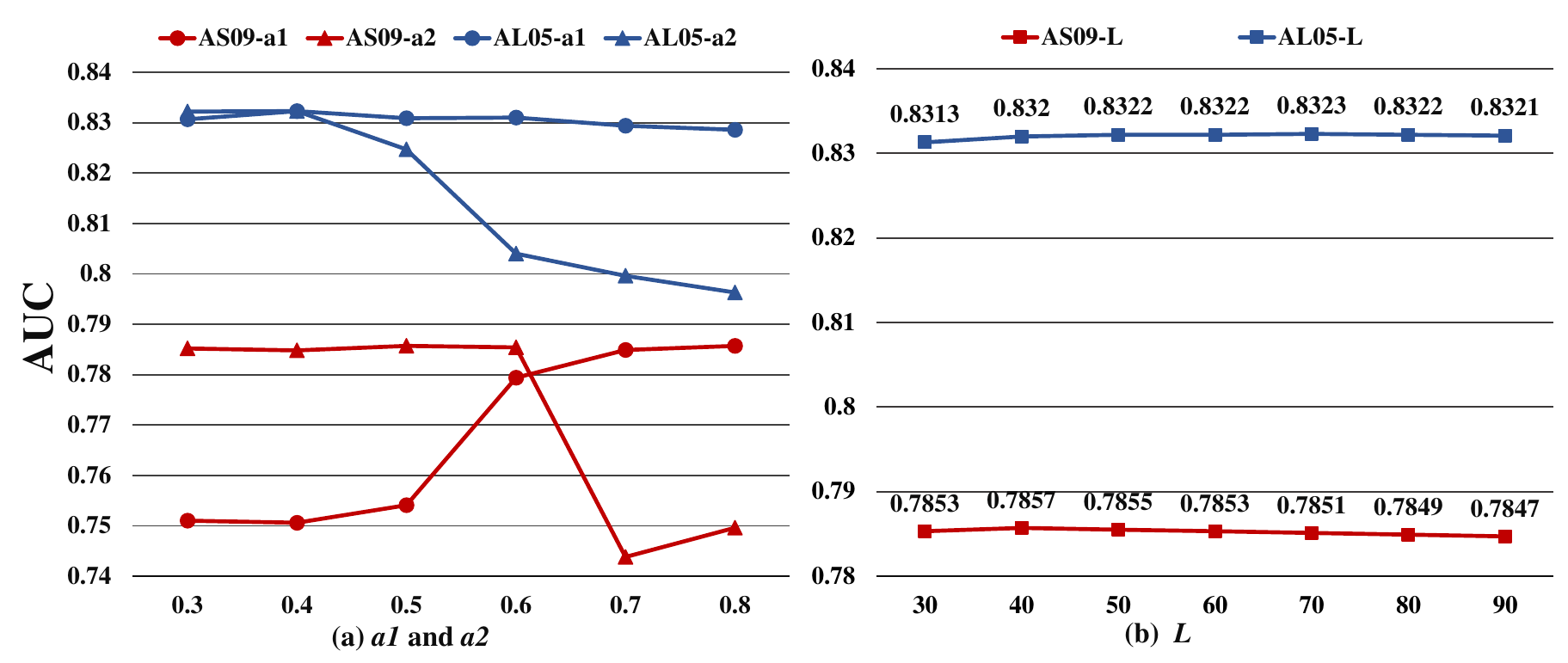}}
\caption{The impact of hyperparameters $a1$, $a2$ and $L$ on model performance. For the AS09, the optimal combination is $\{a1=0.8,a2=0.5\}$; for the AL05, it is $\{a1=0.4,a2=0.4\}$.}
\label{fig-impact}
\end{figure}

We further conduct experiments to investigate the impact of hyperparameters on model performance across the AS09 and AL05. 
Hyperparameter $a1$ and $a2$ in the M-RME method are utilized to adjust the proportion of supplementary information derived from exercises. We observe that varying proportions significantly influence the model's performance, and this impact differs across different datasets. We set the range of values for $a1$ and $a2$ to $[0.3, 0.8]$. Given that these two hyperparameters operate synergistically, to isolate and demonstrate their individual effects on model performance, we generate the fig.\ref{fig-impact}(a) by fixing one hyperparameter of the optimal combinations obtained through exploration while shifting the other. As shown in fig.\ref{fig-impact}(a), on the AS09 which has a higher concept-to-exercises ratio ($0.0069$), higher values of $a1$ and moderate values of $a2$ yield better performance; whereas on the AL05 which has a lower concept-to-exercises ratio ($0.0006$), lower values for both $a1$ and $a2$ yield better performance.
In the TCBA method, the impact of the hyperparameter $L$ on model performance also bears some relation to the characteristics of the dataset. On the AS09, which has shorter average interaction sequences ($72.29$), the performance peaks at $L=40$; on the AL05, which has longer average interaction sequences ($186.55$), the performance peaks at $L=70$.

\begin{table}[htbp]
\scriptsize
\caption{AUC results on AS09, AL05. Ablated model performance, and the degradation compared with the original.}
\label{table-2}
\begin{center}
\begin{tabular}{ccccccc}
\hline
\multirow{2}{*}{Dataset} & \multicolumn{6}{c}{Model}\\
\cline{2-7}
&-T&-CL&-M-CL&-T-CL&-T-M-CL&AlignKT \\
\hline
\multirow{2}{*}{AS09}&0.7807&0.7754&0.7469&0.7713&0.7424&\multirow{2}{*}{0.7857} \\
\cline{2-6}
&-0.0050&-0.0100&-0.0388&-0.0144&-0.0433& \\
\hline
\multirow{2}{*}{AL05}&0.8244&0.8256&0.8232&0.8233&0.8181&\multirow{2}{*}{0.8323} \\
\cline{2-6}
&-0.0079&-0.0067&-0.0091&-0.0090&-0.0142& \\
\hline
\end{tabular}
\end{center}
\end{table}

By observing the selection of hyperparameters and their overall impact on model performance across datasets, we discover that our proposed methods effectively adapt to the different characteristics of the datasets, indicating their excellent interpretability.

\subsection{Ablation Study}
To further evaluate the contribution of the proposed methods to performance improvement, we conduct multiple ablation experiments on the AS09 and AL05. We denote AlignKT without the TCBA method as \emph{-T}, without the contrastive learning module as \emph{-CL}, and without the M-RME method as \emph{-M}. It is important to note that, since the CL module involves on M-RME, we do not remove the M-RME method alone, but instead remove both the M-RME method and the CL module simultaneously. 
Only in the double-module ablation experiments that remove the CL module, we are able to compare the loss incurred by removing the M-RME method. Table \ref{table-2} presents the performance degradation resulting from the removal of relevant modules, and we have the following observations: 
(1) in comparing \emph{AlignKT-CL} with \emph{AlignKT-T}, the loss incurred by removing CL module ($-0.01$) is greater than that by removing TCBA method ($-0.005$) on the AS09. It can be stated that the performance contribution of these innovations on the AS09 is: CL module $>$ TCBA method. While on the AL05, the loss incurred by removing TCBA method ($-0.0079$) is greater than that by removing CL module ($-0.0067$), resulting in an opposite conclusion that CL module $<$ TCBA method. 
(2) In comparing \emph{AlignKT-M-CL} with \emph{AlignKT-CL}, the loss incurred by removing M-RME method on the AS09 ($\Delta=-0.0288$) is significantly greater than that on the AL05 ($\Delta=-0.0024$); while in comparing \emph{AlignKT-T-M-CL} with \emph{AlignKT-T-CL}, the loss incurred by removing M-RME method on the AS09 ($\Delta=-0.0289$) is significantly greater than that on the AL05 ($\Delta=-0.0052$) too. 
\begin{figure}[htbp]
\centerline{\includegraphics[width=\linewidth]{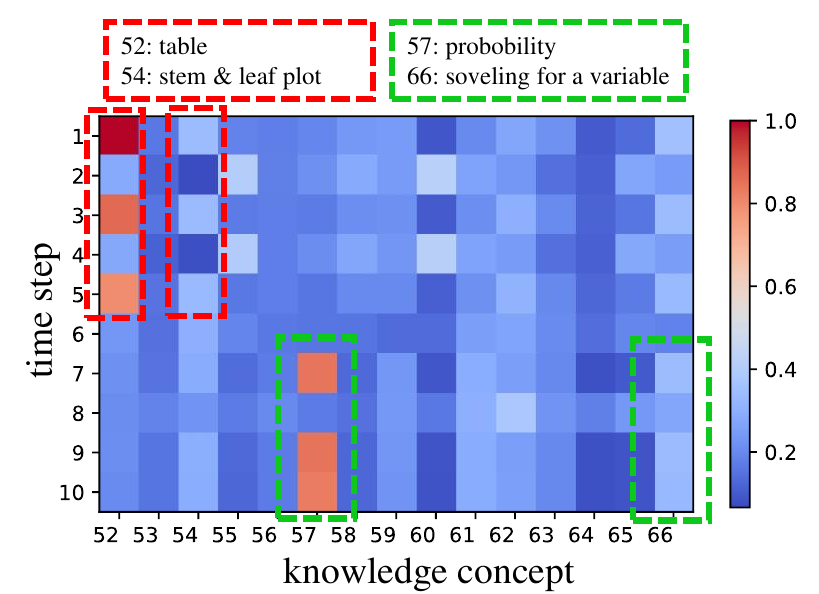}}
\caption{The visualization of a learner's partial knowledge state. The x-axis denotes knowledge concepts, and y-axis denotes time steps. Each row represents the knowledge state at the current time step, with colors closer to the top indicating a higher mastery level of the corresponding knowledge concept.}
\label{fig-visual}
\end{figure}
The results indicate that the M-RME method contributes more significantly to model performance on the AS09 compared to the AL05.

Based on the above two observations, the M-RME method and the CL module, which are designed to address data sparsity and fluctuations in knowledge state, demonstrate more significant improvements on the AS09, where these issues are more prominent (characterized by a higher concept-to-exercise ratio and shorter average sequence length). 
Conversely, the TCBA method, designed to model learner forgetting behavior, demonstrates more significant improvements on the AL05, where this issue is more prominent (characterized by longer average sequence length). These conclusions substantiate that the proposed innovations effectively accomplish our research objectives.

\subsection{Visualization for Interpretability}
In previous works based on attention mechanism, models typically utilize the self-attention weight matrix as a visualization for the interpretability of DLKT. However, this approach is constrained to displaying the relevance of knowledge concepts within a local interaction sequence, but can not reflect learners' mastery level of all knowledge concepts at a global perspective. This is inherently determined by their architecture, and is also a manifestation of the overemphasis on fitting data pattern rather than modeling the knowledge state.

Fig.\ref{fig-visual} visualizes the knowledge state obtained by the AlignKT backend.
At time step $1$, $3$, and $5$, the learner correctly answered exercises related to the concept $52$, resulting in noticeably high indicator values in the corresponding $52$nd column, which signify a high mastery level. Due to the similarity between concept $52$ and concept $54$, a similar pattern can also be observed in the $54$th column. This situation also occurred at time step $7$, $9$, and $10$.

\section{Conclusion}
In this paper, we propose AlignKT, a novel model that introduces a new training paradigm to explicitly model the knowledge state, which addresses the interpretability limitations of previous approaches. By aligning the preliminary knowledge state with an additional interpretable criterion, AlignKT provides a more intuitive representation of the knowledge state, allowing ITS or instructors to more easily assess learners’ mastery levels. Experimental results demonstrate that AlignKT achieves state-of-the-art prediction performance on two out of three real-world datasets, outperforming seven KT baselines. These results validate the success of our approach in explicitly modeling stable knowledge states and improving prediction accuracy, fulfilling the research objectives of improving both interpretability and performance. In the future, we will adapt frontend-to-backend architecture to more latest KT models, driving forward the advancement of interpretability research within the field of knowledge tracing.

\bibliographystyle{IEEEbib}
\bibliography{icme2025references}

\end{document}